\documentclass{article}

\usepackage[utf8]{inputenc} 
\usepackage[T1]{fontenc}    
\usepackage[hyperfootnotes=false]{hyperref}       
\usepackage{url}            
\usepackage{booktabs}       
\usepackage{amsfonts}       
\usepackage{nicefrac}       
\usepackage{microtype}      
\usepackage[numbers]{natbib}
\usepackage{graphicx}
\usepackage{subfigure}
\usepackage{algorithm}
\usepackage{algorithmic}
\usepackage{caption}
\usepackage{wrapfig}
\usepackage{soul}
\usepackage{amsmath, bm}
\usepackage[title]{appendix}
\usepackage{amsthm}
\usepackage[dvipsnames]{xcolor}
\usepackage[final]{neurips_2020}

\newtheorem{prop}{Proposition}
\newtheorem{cor}{Corollary}

\title{Bridging the Gap between Sample-based and One-shot Neural Architecture Search with BONAS}

\author{%
  Han Shi\textsuperscript{\rm 1}\thanks{Equal contribution.}, Renjie Pi\textsuperscript{\rm 2}\footnotemark[1], Hang Xu\textsuperscript{\rm 3}, Zhenguo Li\textsuperscript{\rm 3}, James T. Kwok\textsuperscript{\rm 1}, Tong Zhang\textsuperscript{\rm 1} \\
  \textsuperscript{\rm 1}Hong Kong University of Science and Technology, Hong Kong\\
  \texttt{\{hshiac,jamesk\}@cse.ust.hk, tongzhang@ust.hk} \\
  \textsuperscript{\rm 2}The University of Hong Kong, Hong Kong \\
  \texttt{pipilu@hku.hk} \\
  \textsuperscript{\rm 3}Huawei Noah’s Ark Lab \\
  \texttt{\{xu.hang,li.zhenguo\}@huawei.com} \\
}

\begin{document}

\maketitle
\begin{abstract}
  Neural Architecture Search (NAS) has shown great potentials in finding  better neural network designs. Sample-based NAS is the most reliable approach which aims at exploring the search space and evaluating the most promising architectures. However, it is computationally very costly.  As a remedy, the one-shot approach has emerged as a popular technique for accelerating NAS using weight-sharing. However, due to the weight-sharing of vastly different networks, the one-shot approach is less reliable than the sample-based approach. In this work, we propose BONAS (Bayesian Optimized Neural Architecture Search), a sample-based NAS framework which is accelerated using weight-sharing to evaluate multiple related architectures simultaneously.
  Specifically, we apply a Graph Convolutional Network predictor as surrogate
  model for Bayesian Optimization to select multiple related candidate models in
  each iteration. We then apply weight-sharing to train multiple candidate models
  simultaneously. This approach not only accelerates the traditional sample-based
  approach significantly, but also keeps its reliability. This is because
  weight-sharing among related architectures is more reliable than that in the one-shot approach.
  Extensive experiments are conducted to verify the effectiveness of our method
  over competing algorithms.\footnote[1]{The code is available at \url{https://github.com/pipilurj/BONAS}.}
\end{abstract}

\section{Introduction}
\label{sec:intro}

Designing an appropriate deep
network architecture for each task and data set
is tedious and time-consuming.
Neural architecture search (NAS) \citep{zoph2018learning}, which attempts to
find this architecture automatically,
has aroused significant
interest recently.
Results competitive with hand-crafted architectures have been obtained in
many application areas, such as
natural language processing \citep{luong2018exploring,so2019evolved} and
computer vision
\citep{real2019regularized,ghiasi2019fpn,chen2019detnas,liu2019auto,chu2019fast}.

Optimization in NAS is difficult because
the search space can contain billions of network architectures.
Moreover, the performance (e.g., accuracy)
of a particular
architecture
is computationally expensive to evaluate.
Hence, a central component in NAS
is the
strategy
to search
such a huge space of architectures.
These
strategies
can be broadly categorized into two groups.
Sample-based algorithms
\citep{zoph2016neural,liu2018progressive,real2019regularized,luo2018neural}, which
perform NAS in two phases: (i) search for
candidate architectures with
potentially good performance;
and (ii) query their actual
performance
by full training.
The second category contains one-shot NAS algorithms, which combine architectures in the whole search space together using
weight sharing \citep{pham2018efficient}
or continuous
relaxation \citep{liu2018darts,luo2018neural}
for faster evaluation.
Despite their appealing speed,
one-shot algorithms
suffer from the following: (i) The obtained result can be sensitive to
initialization, which hinders reproducibility; (ii) Constraints need to be
imposed on the search space so as to constrain the super-network size (otherwise,
it may be too large to fit in the memory).
As opposed to one-shot algorithms, sample-based approaches are more flexible with respect to the search space, and can usually find promising architectures regardless of initialization. However, the heavy computation required by sample-based methods inevitably becomes the major obstacle.
In this paper, we aim to develop a more efficient sample-based NAS algorithm while
taking advantage of the weight-sharing paradigm.

Due to the large sizes of most search spaces,
searching for competitive architectures can be very difficult.
To alleviate this issue, Bayesian optimization (BO)
\citep{mockus1978application},
which explicitly considers exploitation and exploration,
comes in handy as an efficient model for search and optimization
problems.
In BO,
a commonly used surrogate model
is the Gaussian process (GP) \citep{snoek2012practical}.
However, its time complexity
increases cubically with the number of samples \citep{snoek2015scalable}.
Hence, it is costly for use in NAS due to the huge search space.
Another drawback of GP in NAS is that it requires a manually designed kernel on
architectures.
While
two
heuristically-designed
kernels
are provided
in
\citep{jin2019auto}
and
\citep{kandasamy2018neural},
they
can neither be easily adapted to different architecture
families nor be further
optimized based on data.
It is still an open issue on how to define a good neural
architecture kernel.
On the other hand, in the query phase,
the traditional approach of fully training
the neural architectures
is costly.
Although early
stopping can be adopted \citep{wang2019sample,white2019bananas}, it cannot reduce
the training time substantially while inevitably compromising the fidelity of the obtained results.
In one-shot methods,
weight-sharing
is performed on the whole space of
sub-networks
\citep{pham2018efficient}.
These sub-networks can be very different and so
sharing
their weights may not be a good idea.

\begin{figure}
\includegraphics[width=\columnwidth]{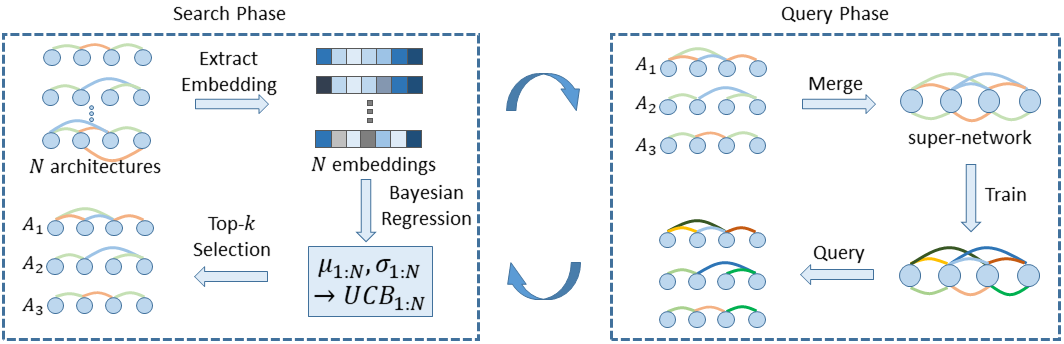}
\caption{Overview of the proposed BONAS. In the search phase, we use GCN embedding extractor and Bayesian Sigmoid Regression as the surrogate model for Bayesian Optimization and multiple candidate architectures are selected. In the query phase, we merge them as a super-network. Based on the trained super-network, we can query each sub-network using corresponding paths.
\label{fig:framework}}
\end{figure}

To alleviate these problems, we present BONAS (Bayesian Optimized Neural Architecture Search), which is a sample-based NAS algorithm combined with weight-sharing
(Figure~\ref{fig:framework}).
In the search phase, we first
use a graph convolutional network (GCN) \citep{kipf2016semi}
to produce embeddings for the
neural architectures.
This naturally handles the graph structures of neural architectures, and
avoids defining GP's kernel function.
Together with a novel
Bayesian sigmoid regressor, it
replaces the GP
as
BO's
surrogate model.
In the query phase, we construct a super-network from a batch of promising candidate
architectures, and train them by uniform sampling.
These candidates are then
queried
simultaneously based on the learned weight of the super-network.
As weight-sharing is now performed only on a small subset of similarly-performing sub-networks
with high BO scores,
this is more reasonable
than
sharing the weights of all sub-networks in the search space
as in one-shot NAS methods
\citep{pham2018efficient}.

Empirically, the proposed BONAS outperforms state-of-the-art
methods.
We observe consistent gains on multiple
search spaces for vision and NLP tasks. These include the standard benchmark data
sets of
NAS-Bench-101 \citep{ying2019bench} and NAS-Bench-201 \citep{dong2020bench}
on convolutional  architectures,
and
a new NAS benchmark data set
LSTM-12K
we recently collected
for LSTMs.
The proposed algorithm also finds
competitive models efficiently in open-domain search with the NASNet search space \citep{zoph2018learning}.

The contributions of this paper are as follows.
    (i) We improve the efficiency of sample-based NAS using Bayesian optimization in combination with a novel GCN embedding extractor and Bayesian Sigmoid Regression to select candidate architectures.
    (ii)  We accelerate the evaluation of sample-based NAS by training multiple related architectures simultaneously using weight-sharing.
    (iii) Extensive experiments on both closed and open domains demonstrate the efficiency of the proposed method. BONAS achieves consistent gains on different benchmarks compared with competing baselines.
It bridges the gap between training speeds of sample-based and one-shot NAS methods.

\section{Related Work}
\subsection{Bayesian Optimization}
\label{sec:bo}

Bayesian optimization (BO)
\citep{mockus1978application},
with the Gaussian process (GP) \cite{snoek2012practical} as the
underlying surrogate model,
is a popular technique for finding the globally optimal solution
of an optimization problem.
To guide the search,
an acquisition function is used
to balance exploitation and exploration \citep{shahriari2015taking}.
Common examples include the maximum probability of improvement (MPI) \cite{kushner1963new}, expected improvement (EI) \cite{mockus1978application} and
upper confidence bound (UCB) \cite{srinivas2009gaussian}.
In this paper, we focus on the UCB,
whose
exploitation-exploration tradeoff is explicit and easy to adjust.
Let the hyperparameters of BO's surrogate model be $\Theta$, and the observed data be
$\mathcal{D}$.
The UCB for a new sample $x$
is:
\begin{equation}
a_{\text{UCB}}(x;\mathcal{D},\Theta) = \mu(x;\mathcal{D},\Theta)+\gamma\sigma(x;\mathcal{D},\Theta),\label{eq:EI}
\end{equation}
where
$\mu(x;\mathcal{D},\Theta)$ is the predictive mean of the output from the surrogate
model, $\sigma^{2}(x;\mathcal{D},\Theta)$ is the corresponding predictive variance,
and $\gamma>0$ is a tradeoff parameter.
A
larger
$\gamma$ puts
more emphasis on exploration, and vice versa.

\subsection{BO for Neural Architecture Search}
Recently,  BO is also used in NAS \cite{jin2019auto,kandasamy2018neural}.
Its generic procedure is shown in Algorithm~\ref{alg:bo-nas}.
Since the NAS search $\mathcal{A}$ is huge,
each BO iteration typically only considers a pool of architectures, which
is generated, for example, by an evolutionary algorithm (EA) \citep{real2019regularized}.
An acquisition function score is computed for each architecture in the pool,
and architectures with the top scores
are then selected
for query. The procedure is repeated until convergence.




\begin{minipage}[t]{0.48\textwidth}
\begin{algorithm}[H]
\caption{Generic BO procedure for NAS.}
\label{alg:bo-nas}
\begin{algorithmic}[1]
\STATE{randomly select $m_0$ architectures $\mathcal{D}$ from search
space
$\mathcal{A}$
for full training;}
\STATE{initialize surrogate model using
$\mathcal{D}$;}
\REPEAT
\STATE{sample candidate pool $\mathcal{C}$ from $\mathcal{{A}}$;}
\FOR{each candidate $m$ in $\mathcal{C}$}
\STATE{score $m$ using acquisition function;}
\ENDFOR
\STATE{$M \leftarrow$ candidate(s) with the top score(s);}
\STATE{(query): obtain actual performance of $M$;}
\STATE{add $M$ and its performance to $\mathcal{D}$;}
\STATE{update surrogate model with the enlarged $\mathcal{D}$;}
\UNTIL{convergence.}
\end{algorithmic}
\end{algorithm}
\end{minipage}
\hfill
\begin{minipage}[t]{0.51\textwidth}
\begin{algorithm}[H]
\caption{BONAS.}
\label{alg:BONAS-Search-Procedure}
\begin{algorithmic}[1]
\STATE{randomly select $m_0$ architectures $\mathcal{D}$ from
search space $\mathcal{A}$ for weight-sharing training;}
\STATE{initialize GCN and BSR using $\mathcal{D}$;}
\REPEAT
\STATE{sample candidate pool $\mathcal{C}$ from $\mathcal{{A}}$ by EA;}
\FOR{each candidate $m$ in
$\mathcal{C}$}
\STATE{embed $m$ using GCN;}
\STATE{compute mean and variance
using BSR;}
\STATE{compute UCB in (\ref{eq:EI});}
\ENDFOR
\STATE{$M \leftarrow$ candidates with the top-$k$ scores;}
\STATE{(query): train $M$ with weight-sharing;}
\STATE{add $M$ and their performances to $\mathcal{D}$;}
\STATE{update GCN and BSR with the enlarged $\mathcal{D}$;}
\UNTIL{convergence.}
\end{algorithmic}
\end{algorithm}
\end{minipage}

\section{Proposed Method}
\label{sec:method}

As can be seen from
Algorithm~\ref{alg:bo-nas},
the key issues for the successful application of BO for NAS are:
(i) How to represent an architecture?
(ii) How to find good candidates with the surrogate model?
In particular, candidates with high
acquisition scores should have high actual performance; (iii) How to
query the selected candidates efficiently?

In this section,
we design a surrogate model combining a GCN embedding extractor and a Bayesian sigmoid regressor
(Section~\ref{sec:sur}).
To alleviate the cost of full training in each query, we adopt the weight-sharing
paradigm to query a batch of promising architectures together (Section~\ref{subsec:weight-share}).
Figure~\ref{fig:framework} shows an overview of the proposed algorithm (Section~\ref{subsec:Search-with-Alternate}).

\subsection{Finding Potential Candidates with the Surrogate Model}
\label{sec:sur}

In this section, we introduce a surrogate model which consists of a GCN embedding extractor and a Bayesian sigmoid regressor.

\subsubsection{Representing Neural Networks using GCN}
\label{sec:gcn}

\begin{wrapfigure}{r}{9cm}
\includegraphics[width=0.59\columnwidth]{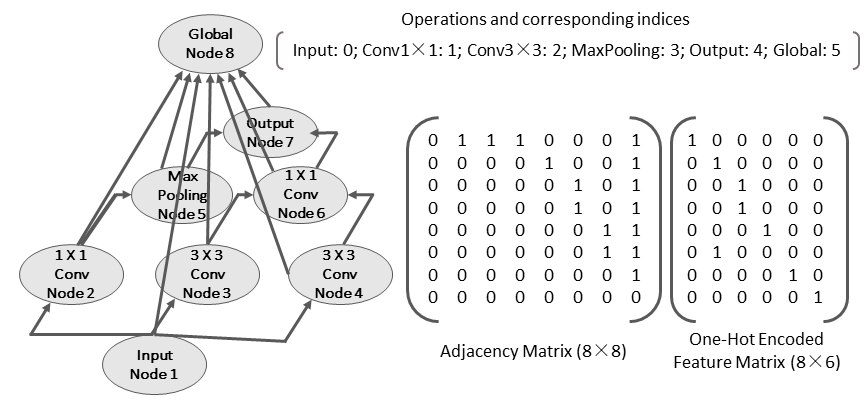}
\caption{Encoding of an example cell in NAS-Bench-101. \label{fig:cell_emb}}
\end{wrapfigure}

A neural network can be represented as a directed attributed graph. Each node represents an operation
(such as a $1\times 1$ convolution in CNN, and ReLU activation in LSTM) while edges represent data flows \citep{zhang2019graph}.
Figure~\ref{fig:cell_emb} shows an example on NAS-Bench-101.
Since a NAS-Bench-101 architecture is obtained by stacking multiple repeated cells, we only consider the embedding of such a cell.
Graph connectivity is encoded by the
adjacency matrix $\boldsymbol{A}$.
Individual operations are encoded as
one-hot vectors,
and then aggregated to form
the feature matrix $\boldsymbol{X}$.

Existing works often use MLP or LSTM to encode networks
\citep{liu2018progressive,luo2018neural,wang2019alphax}. However,
it is more natural to use GCN,
which can well preserve the graph's structural information \citep{kipf2016semi}.
Besides, a MLP only allows fixed-sized inputs, while  a
GCN can handle input graphs with variable numbers of nodes.

The standard GCN is only used to produce node embeddings \citep{kipf2016semi},
while here the target is to obtain an
embedding for the whole
graph.
To solve this problem, following \citep{scarselli2008graph},
we connect all nodes in the graph to an additional ``global" node
(Figure~\ref{fig:cell_emb}).  The one-hot encoding scheme is also extended to include these new connections as a
new operation. The embedding of the global node
is then used as the
embedding of the whole graph.

To train the GCN, we feed its output
(cell embedding)
to a regressor for
accuracy prediction.
In the experiments,
we use a single-hidden-layer network with sigmoid function, which
constrains the prediction to be in $[0, 1]$ (it can
be easily scaled
to a different range
when another performance
metric is used).
This regressor is then trained
end-to-end
with the GCN
by minimizing the square loss.


BANANAS \citep{white2019bananas}, an independent concurrent work with this paper,
also tries to encode the graph structure by a so-called {\em path encoding}
scheme, which is then fed to a MLP (called {\em meta
neural network})
for performance estimation.
The path encoding scheme
is similar to the bag-of-words representation for documents, and its combined use
with a simple MLP is
less powerful than the GCN (as will be demonstrated empirically in Section~\ref{subsec:few}).
Moreover, its encoding vector scales exponentially in size with
the number of nodes, and so may not be scalable to large cells. In
\citep{white2019bananas}, they need to truncate the encoding by eliminating
paths that are less likely.


\subsubsection{Bayesian Sigmoid Regression}
\label{sec:bsr}

To compute the mean and variance of the architecture's performance in
(\ref{eq:EI}),
we introduce a Bayesian sigmoid regression (BSR) model. This is
inspired by the Bayesian linear regression (BLR) in neural networks \citep{snoek2015scalable}.
For an architecture (graph) with adjacency matrix $\boldsymbol{A}$ and feature matrix $\boldsymbol{X}$,
let $\boldsymbol{\phi}(\boldsymbol{A},\boldsymbol{X})$ be the learned embedding in Section~\ref{sec:gcn}.
Given a set
$\mathcal{D}$
of
$N$ trained architectures $\{(\boldsymbol{A}_{i},\boldsymbol{X}_{i})\}$ with known performance (accuracy)
values $\{t_{i}\}$,
the corresponding embedding vectors
$\{\boldsymbol{\phi}(\boldsymbol{A}_{i},\boldsymbol{X}_{i})\}$ are
stacked to form a design matrix
$\mathbf{\Phi}$
with $\Phi_{ij}=\phi_{j}(\boldsymbol{A}_{i},\boldsymbol{X}_{i})$.
Recall from
Section~\ref{sec:gcn}
that the final layer of the GCN
predictor
contains a sigmoid function (rather than the linear function in
BLR).
Instead of fitting the true performance $t$,
we estimate the value before the sigmoid, i.e.,
$y=\text{logit}(t)=\log(t/(1-t))$, such that we can convert nonlinear
regression to a linear regression problem. We denote the vector of regression
values by $\boldsymbol{y} = [y_1, y_2, \dots, y_N]^T $.

In Section~\ref{sec:gcn},
the GCN
is trained
with the square loss.
However, when training
BO's surrogate model,
we are more interested
in predicting
high-performance
architectures
as accurately as possible.
Inspired by the focal loss in classification \citep{lin2017focal}, we use the following exponentially weighted loss, which
puts more emphasis on models with higher accuracies:
\begin{equation}
L_{exp}=\frac{1}{N}\sum_{i=1}^{N}(\exp(t_{i})-1)(\widetilde{t_{i}}-t_{i})^{2}, \label{eq:wl}
\end{equation}
where
$\widetilde{t_{i}}$ is
GCN's predicted accuracy on architecture $i$.



For a candidate network $(\boldsymbol{A},\boldsymbol{X})$, BO has to evaluate its acquisition score.
For UCB,
this involves estimating the predictive mean and predictive variance of the
predicted accuracy.
From
\citep{bishop2006pattern},
the predictive mean $\mu$ of $\text{logit}(t)$ is:
\begin{equation}
\mu(\boldsymbol{A},\boldsymbol{X};\mathcal{D},\alpha,\beta) = \boldsymbol{m}_{N}^{T}\boldsymbol{\phi}(\boldsymbol{A},\boldsymbol{X}),\label{eq:mean}
\end{equation}
where
$\boldsymbol{m}_{N}=\beta \boldsymbol{S}_{N}\boldsymbol{\Phi}^{T}\boldsymbol{y}$, $\boldsymbol{S}_{N}
=(\alpha \boldsymbol{I}+\beta\boldsymbol{\Phi}^{T}\boldsymbol{\Phi})^{-1}
$, $\boldsymbol{I}$ is
the identity matrix, and
$(\alpha, \beta)$ are precision parameters  that can be estimated by
maximizing the marginal likelihood \citep{snoek2012practical}.
By considering
only
the weight uncertainty in the last
layer,
the
predictive variance of
$\text{logit}(t)$
is
\citep{bishop2006pattern}:
\begin{equation}
\sigma^{2}(\boldsymbol{A},\boldsymbol{X};\mathcal{D},\alpha,\beta)=
\boldsymbol{\phi}(\boldsymbol{A},\boldsymbol{X})^{T}\boldsymbol{S}_{N}\boldsymbol{\phi}(\boldsymbol{A},\boldsymbol{X})
+
1/\beta.\label{eq:variance}
\end{equation}
In the following, we show how to convert this to the predictive variance of $t$.

First, note that $t$ follows the logit-normal distribution\footnote{The logit-normal distribution is given by: $p(t;\mu, \sigma) = \frac{1}{\sigma\sqrt{2\pi}}\frac{1}{t(1-t)}\exp\left(-\frac{(\text{logit}(t)-\mu)}{2\sigma^2}\right)$.}
\citep{mead1965generalised}.
However, its $\text{E}[t]$ and $\text{var}[t]$
cannot
be analytically computed. To alleviate this problem,
we rewrite
$E[t]$ and $E[t^2]$
as
\[ E[t] =
\smallint \text{sigmoid}(x)\mathcal{N}(x|\mu, \sigma^2)dx, \;\;
E[t^2] =
\smallint (\text{sigmoid}(x))^2\mathcal{N}(x|\mu, \sigma^2)dx, \]
where
$\mathcal{N}(x|\mu, \sigma^2)$ is the normal distribution with mean $\mu$ and variance $\sigma^2$.
Let $\Phi(x)=\int_{-\infty}^x\mathcal{N}(z|0, 1)dz$ be the cumulative distribution function of $\mathcal{N}(x|0, 1)$.
We approximate
$\text{sigmoid}(x)$
with
$\Phi(\lambda x)$ for some
$\lambda$,
and similarly $(\text{sigmoid}(x))^2$ with
$\Phi(\lambda\alpha (x+\beta))$, for some $\lambda,\alpha,\beta$.
With these approximations,
the following Proposition shows that the integrals can be
analytically computed.
Proof is in Appendix~A.

\begin{prop} \label{thm:1}
For given $\alpha$ and $\beta$,
$\int\Phi(\alpha(x+\beta))\mathcal{N}(x|\mu, \sigma^2)dx =
\Phi\left(\frac{\alpha(\mu+\beta)}{(1+\alpha^2\sigma^2)^{1/2}} \right)$.
\end{prop}

\begin{cor} \label{prop:1}
The expectation and variance of the logit-normal distribution can be approximated as:
\begin{eqnarray*}
E[t]
\simeq \text{sigmoid}\left(\frac{\mu}{\sqrt{1+\lambda^2\sigma^2}} \right), \;
var[t]
\simeq \text{sigmoid}\left(\!\frac{\alpha(\mu+\beta)}{\sqrt{1+\lambda^2\alpha^2\sigma^2}}
\!\right)-\left(\!\!\text{sigmoid}\left(\!\frac{\mu}{\sqrt{1+\lambda^2\sigma^2}}
\!\right)\!
\right)^2,
\end{eqnarray*}
where $\lambda^2=\pi/8,
\alpha=4-2\sqrt{2}$ and $\beta=-\log(\sqrt{2}+1)$.
\end{cor}

\begin{figure}[t]
\subfigure[$\sigma=1$.]{\includegraphics[width=.24\columnwidth]{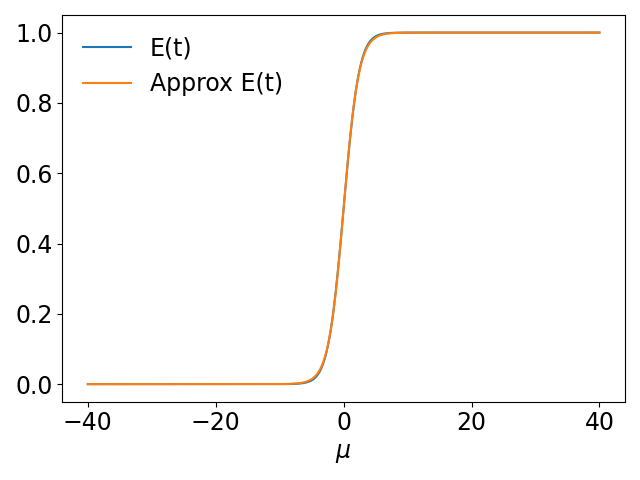}}
\subfigure[$\mu=0$.\label{fig:overlap}]{\includegraphics[width=.24\columnwidth]{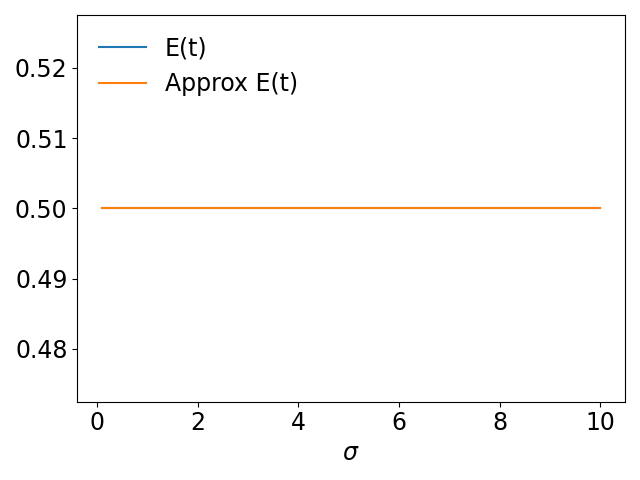}}
\subfigure[$\sigma=1$.]{\includegraphics[width=.24\columnwidth]{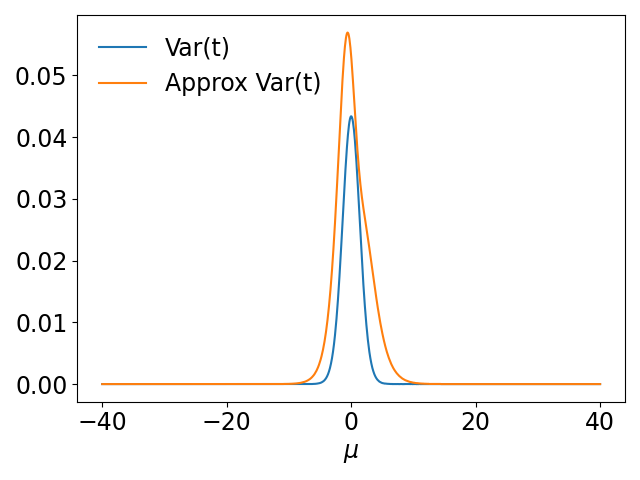}}
\subfigure[$\mu=0$.]{\includegraphics[width=.24\columnwidth]{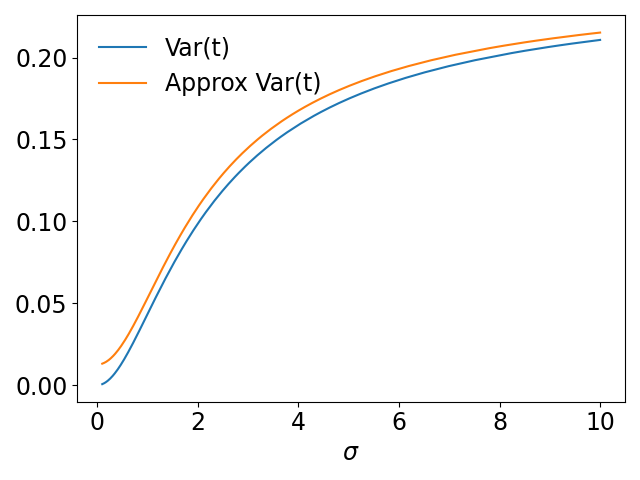}}
\caption{Plots for the true and approximate values of $E[t]$ and $var[t]$.}
\label{fig:approximation}
\end{figure}

Figure~\ref{fig:approximation} shows illustrations of the
approximations.
From the
obtained $\text{E}[t]$ and $\text{var}[t]$,
one can plug into the
UCB score in (\ref{eq:EI}), and
use this to select the next (architecture) sample from the pool.

BANANAS \citep{white2019bananas} also uses BO to search.  To obtain the predictive variance in UCB,
they
compute the empirical variance over
the outputs of
an ensemble
of MLP predictors.
A small ensemble leads to biased estimation, while a large ensemble is
computationally expensive to train and store.
In contrast,
we only use one GCN predictor, and the variance is obtained directly
from BSR.






\subsection{Efficient Estimation of Candidate Performance}
\label{subsec:weight-share}

In sample-based NAS algorithms,
an architecture is selected in each iteration for full training
\cite{real2019regularized,wang2019sample,white2019bananas,zoph2018learning}, and is
computationally expensive.
To alleviate this problem,
we select
in each BO iteration
a batch of $k$ architectures
$\{(\boldsymbol{A}_i,\boldsymbol{X}_i)\}_{i=1}^k$
with the top-$k$ UCB scores,
and then train them together as a super-network by weight-sharing.
While weight-sharing has been commonly used in
one-shot NAS algorithms
\cite{liu2018darts,pham2018efficient, xie2018snas},
their super-networks usually contain architectures from the whole search
space. This makes it infeasible to train each architecture fairly
\cite{chu2019fairnas}, and some architectures
may not be trained as sufficiently as others.
In contrast,
we only use a small number
of
architectures to form the super-network ($k=100$ in the experiments).
With such a small $k$, training time can be allocated to the sub-networks more
evenly.
Moreover, since these $k$ chosen architectures have top UCB scores,
they are promising candidates and
likely to contain common useful structures for the task.
Thus,
weight-sharing
is expected to be more
efficient.

As illustrated in
Figure~\ref{fig:framework}, during the query phase,
we construct the super-network with adjacency matrix
$\hat{\boldsymbol{A}} = \boldsymbol{A}_1||\boldsymbol{A}_2 || \dots ||\boldsymbol{A}_k$, and feature matrix $\hat{\boldsymbol{X}} = \boldsymbol{X}_1||\boldsymbol{X}_2 ||\dots
||\boldsymbol{X}_k$,
where $||$
denotes the logical OR operation.
The super-network $(\hat{\boldsymbol{A}},\hat{\boldsymbol{X}})$
is then trained
by uniformly sampling from the architectures $\{(\boldsymbol{A}_i, \boldsymbol{X}_i)\}$
\citep{chu2019fairnas}. In each iteration, one sub-network $(\boldsymbol{A}_i, \boldsymbol{X}_i)$
is
randomly sampled
from the super-network,
and only the corresponding
(forward and backward propagation)
paths
in it
are activated.
Finally, we evaluate each sub-network by only forwarding data along
the corresponding paths in the super-network.

In ENAS \citep{pham2018efficient},
the operation weights
are reused
along the whole
search process.
Hence, networks evaluated later in the process are trained with longer budgets,
which may render the evaluation unfair.
In the proposed algorithm, we
reinitialize
the operation weights
at each query phase,
ensuring
that each sub-network is trained for the same number of iterations.

\subsection{Algorithm BONAS}
\label{subsec:Search-with-Alternate}
The whole procedure, which will be called BONAS (Bayesian Optimized Neural Architecture Search), is shown in Algorithm~\ref{alg:BONAS-Search-Procedure}.
Given the search space $\mathcal{{A}}$,
we start with
a set $\mathcal{D}$ of
$m_0$
random
architectures
$\{(\boldsymbol{A}_{i},\boldsymbol{X}_{i})\}$, which have been
queried and the corresponding
performance values
$\{t_{i}\}$ known.
The GCN embedding extractor and BSR are then
trained
using $\mathcal{D}$ (Section~\ref{sec:sur}).

In each search iteration, a pool $\mathcal{C}$ of candidates are sampled from
$\mathcal{{A}}$ by evolutionary algorithm (EA).
For each candidate, its embedding is generated by the GCN, which is used by BSR
to compute the mean and variance of its predicted accuracy. The UCB score is then
obtained from (\ref{eq:EI}).
Candidates with the top-$k$ UCB scores are selected and queried
using weight sharing (Section~\ref{subsec:weight-share}).
The evaluated models and their performance values are added to $\mathcal{D}$.
The GCN predictor and BSR
are then updated using the enlarged $\mathcal{D}$.
The procedure is repeated until convergence.

\section{Experiments}
In the following experiments,
we use NAS-Bench-101 \citep{ying2019bench},
which is the largest NAS benchmark data set (with 423K convolutional
architectures),
and
the more recent
NAS-Bench-201
\citep{dong2020bench},
which uses a different search space (with 15K architectures)
and is applicable to almost any NAS algorithm.

As both NAS-Bench-101 and NAS-Bench-201 focus on
convolutional architectures,
we also construct another
benchmark data set
(denoted
LSTM-12K),
containing 12K LSTM models
trained on the Penn TreeBank data set
\citep{marcus1993building} following the same setting
in ENAS \citep{pham2018efficient}.
Each LSTM cell,
with an adjacency matrix and a list of operations,
is represented by a string.
In the search space, there are $4$ possible activation functions (tanh, ReLU, identity, and
sigmoid) and each node takes one previous node as input. The architecture is
obtained by selecting the activation functions and node
connections.
Due to limitation on computational resources, we only
sample architectures with
$8$
or fewer
nodes.
We randomly sampled
12K cell structures.
The perplexity is used as the metric to evaluate performances of the models.
More details on
the training setup and
search space are
in Appendix~B.

Experiments are also performed in the
open-domain scenario using the NASNet search space
\citep{zoph2018learning}.
All
experiments are performed on NVIDIA Tesla V100 GPUs.

\subsection{Comparison of Predictor Performance\label{subsec:few}}

In this section, we
demonstrate superiority of the proposed GCN predictor over existing MLP and
LSTM predictors
in \citep{wang2019alphax}, and the meta NN in \citep{white2019bananas}.
The GCN
has four hidden layers
with $64$ units each.
Training
is performed by minimizing
the square loss,
using the Adam optimizer
\citep{kingma2014adam} with a learning rate of 0.001 and a mini-batch
size of $128$.
For the MLP predictor,
we follow \citep{wang2019alphax} and use
5 fully-connected layers, with 512, 2048, 2048, 512 and 1
units, respectively.
As for the LSTM predictor, the sizes of both the hidden layer and embedding are 100. The last LSTM
hidden layer is connected to a fully-connected layer.
For the meta NN, we use an
ensemble of $3$ predictors (each being a
fully-connected neural network with
$10$ layers, and
$20$ hidden units in each layer) and apply the full-path encoding scheme.

\begin{wraptable}{r}{7cm}
\caption{Correlation between the model's predicted and actual performance.}
\resizebox{0.5\textwidth}{!}{
\begin{tabular}{cccc}
\hline
 & NAS-Bench-101 & NAS-Bench-201 & LSTM-12K\tabularnewline
\hline
MLP & 0.830 & 0.865 & 0.530\tabularnewline
LSTM & 0.741 & 0.795 & 0.560\tabularnewline
Meta NN & 0.648 & 0.967 & 0.582 \tabularnewline
GCN & \textbf{0.841} & \textbf{0.973} & \textbf{0.742}\tabularnewline
\hline
\end{tabular}}
\label{tab:predictor}
\end{wraptable}

Experiments are performed on the NAS-Bench-101,
NAS-Bench-201, and
LSTM-12K data sets.
For each data set,
we use $85\%$ of the data for training, $10\%$ for validation, and the rest for testing.
For performance evaluation,
as in \citep{wang2019alphax},
we
use the correlation coefficient between the model's predicted and actual
performance values (i.e., testing accuracy
on NAS-Bench-101 and NAS-Bench-201, and
perplexity
on LSTM-12K).
Table~\ref{tab:predictor}
shows the results.
As can be seen, the GCN predicts the performance
more accurately
than the other three predictors.


\begin{figure}[t]
\centering
\subfigure{\includegraphics[width=0.32\columnwidth]{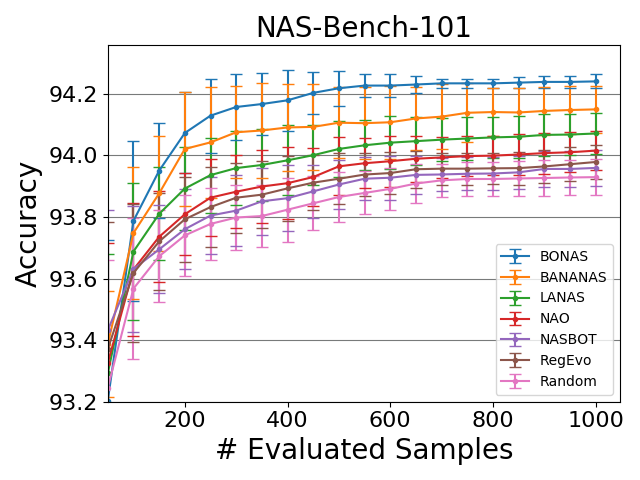}}
\subfigure{\includegraphics[width=0.32\columnwidth]{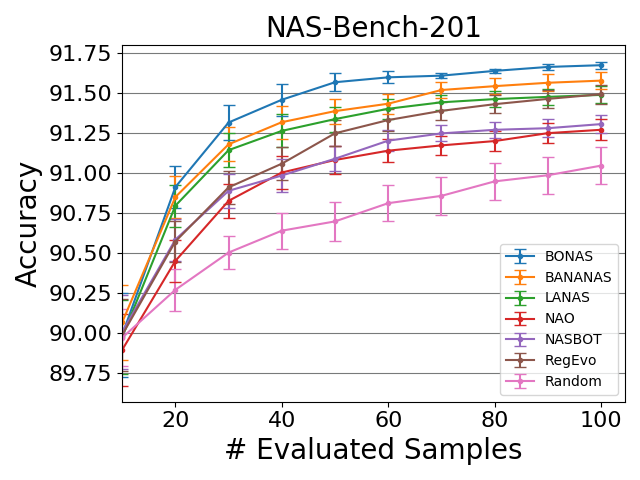}}
\subfigure{\includegraphics[width=0.32\columnwidth]{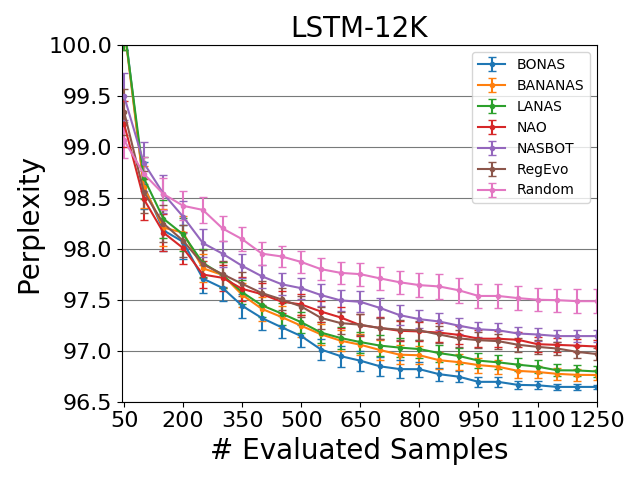}}
\caption{Performance (accuracy/perplexity) of the best model versus number of evaluated samples.}
\label{fig:Current-best-accuracy}
\end{figure}

\subsection{Closed Domain Search\label{subsec:single}}

In this section, we study the search efficiency of the proposed BONAS algorithm.
In step~1 of Algorithm~\ref{alg:BONAS-Search-Procedure},
we start with $10$ random architectures that are fully trained.\footnote{These 10
are counted towards the total number of
architectures
sampled.}
In step~4,
for NAS-Bench-201, the adjacency matrix is fixed and not mutated by the EA
sampler.
The candidate pool $\mathcal{C}$ is of size 10,000 and
$\gamma= 0.5$
in (\ref{eq:EI}).
As the benchmarks already contain the performances of all models, we query each
architecture ($k=1$) by directly obtaining its accuracy from the data sets
rather than using the weight-sharing query scheme in Section~\ref{subsec:weight-share}.
Since only
the search phase
but not the query phase is
performed,
this allows us to demonstrate the search efficiency of BONAS more clearly.

Note that one-shot methods do not search for models iteratively.
We compare
BONAS
with the following state-of-the-art sample-based
NAS baselines:
(i) Random search \citep{yang2019evaluation}, which explores the search space randomly
without exploitation; (ii) Regularized evolution \citep{real2019regularized},
which uses a heuristic evolution process for exploitation;
(iii)
NASBOT
\citep{kandasamy2018neural}, which uses BO with a manually-defined
kernel on architectures; (iv) Neural Architecture Optimization (NAO) \citep{luo2018neural}, which finds the architecture in a continuous embedding space with gradient descent;
(v) LaNAS \citep{wang2019sample},
which estimates the architecture performance in a
coarse model
subspace by Monte
Carlo tree search and (vi) BANANAS \cite{white2019bananas}, which applies a traditional BO
framework for the NAS problem.
The experiment is repeated 50 times, and the averaged result with standard deviation reported.

Following \citep{wang2019sample,wang2019alphax},
Figure~\ref{fig:Current-best-accuracy} shows the performance of the best model
after using a given number of architecture samples.
As can be seen, BONAS
consistently
outperforms the other search algorithms.

\subsection{Open Domain Search \label{subsec:Open-domain-Multi-objective}}

In this section, we perform NAS on  the
NASNet search space \citep{zoph2018learning}
using the CIFAR-10 data set. Following \citep{liu2018darts}, we allow $4$ blocks inside a cell.
In step~10 of Algorithm~\ref{alg:BONAS-Search-Procedure},
$k=100$ models
are merged
to a super-network and trained for $100$ epochs using the procedure discussed in Section~\ref{subsec:weight-share}.
In each epoch,
every sub-network is trained for the same number of  iterations.
The other experimental settings are the same as in
Section~\ref{subsec:single}.

The proposed BONAS algorithm
is compared with the following state-of-the-art sample-based NAS
algorithms (results of the various baselines are taken from the corresponding
papers):
(i) NASNet
\citep{zoph2018learning},
which uses reinforcement learning to sample architectures directly;
(ii) AmoebaNet
\citep{real2019regularized}, which finds the architecture by an evolution
algorithm;
(iii) PNASNet
\citep{liu2018progressive},
which searches the
architecture progressively combined with a predictor;
(iv) NAO \citep{luo2018neural}; (v) LaNet
\citep{wang2019sample};
(vi) BANANAS.
We also show results of the one-shot NAS methods, including: (i) ENAS
\citep{pham2018efficient},
which finds
the model by parameter sharing;
(ii) DARTS
\citep{liu2018darts},
which applies
continuous relaxation for super-network training;
(iii)
BayesNAS \citep{zhou2019bayesnas},
which considers the dependency on architectures;
and (iv) ASNG-NAS
\citep{akimoto2019adaptive},
which proposes a stochastic natural gradient method for the NAS problem.

\begin{table}[t]
\begin{center}
\caption{Performance of open-domain search on CIFAR-10. \#blocks is the number of
blocks in a cell, and cutout
\citep{devries2017improved}
is a popular data augmentation strategy
in NAS.
For one-shot NAS methods, the samples are not explored one by one, and
the number of samples evaluated
is marked "-".}
\label{tab:Comparison-of-different}

\centering{}
\resizebox{\textwidth}{!}{
\begin{tabular}{cccccc}
\hline
& \#blocks & \#params & top-1 err (\%) & \#samples evaluated & GPU days\tabularnewline
\hline
GHN+cutout \citep{zhang2019graph} & 7 & 5.7 M &  2.84 & - &0.84 \tabularnewline
LaNet+cutout \citep{wang2019sample} & 7 & 3.2 M & 2.53 & 803 & 150\tabularnewline\hline
ASNG-NAS+cutout \citep{akimoto2019adaptive} & 5 & 3.9 M & 2.83 & - & 0.11\tabularnewline
ENAS+cutout \citep{pham2018efficient} & 5 & 4.6 M & 2.89 & - &
0.45\tabularnewline
NASNet-A+cutout \citep{zoph2018learning} & 5 & 3.3 M & 2.65 & 20,000 & 2,000\tabularnewline
AmoebaNet-B+cutout \citep{real2019regularized} & 5 & 2.8 M & 2.55 & 27,000 & 3,150\tabularnewline
NAO \citep{luo2018neural} & 5 & 10.6 M & 3.18 & 1,000 & 200 \tabularnewline
\hline
DARTS+cutout \citep{liu2018darts} & 4 & 3.3 M & 2.76 & - & 1.5\tabularnewline
BayesNAS+cutout \citep{zhou2019bayesnas} & 4 & 3.4 M & 2.81 & - & 0.2\tabularnewline
PNASNet-5 \citep{liu2018progressive} & 4 & 3.2 M & 3.41 & 1,160 & 225\tabularnewline
BANANAS+cutout \citep{white2019bananas} & 4 & 3.6 M & 2.64 & 100 & 11.8 \tabularnewline
BONAS-A+cutout & 4 & 3.45 M & 2.69 & 1,200 & 2.5\tabularnewline
BONAS-B+cutout & 4 & 3.06 M & 2.54 & 2,400 & 5.0\tabularnewline
BONAS-C+cutout & 4 & 3.48 M & 2.46 & 3,600 & 7.5\tabularnewline
BONAS-D+cutout & 4 & 3.30 M & \textbf{2.43} & 4,800 & 10.0\tabularnewline
\hline
\end{tabular}}
\end{center}
\end{table}
Results on CIFAR10 are shown in Table~\ref{tab:Comparison-of-different}.
We list 4 BONAS models (A, B, C, D) obtained with different numbers of
evaluated samples.
Note that different papers may use different numbers of blocks in the experiment, and comparison across different
search spaces
may not be fair.
As can be seen
from Table~\ref{tab:Comparison-of-different},
BONAS outperforms all the other algorithms
in terms of the top-$1$ error.
Moreover,
by using weight-sharing query,
BONAS
is very efficient compared with the other sample-based NAS algorithms.
For example, BONAS can sample and query $4800$ models in around $10$ GPU days,
while BANANAS only queries $100$ models in $11.8$ GPU days.
We show the search progress of BONAS in Appendix~C, and
example architectures learned by BONAS  are
in Appendix~D.


\subsection{Transfer Learning}
\label{sec:dataset}

As in \citep{liu2018darts,luo2018neural,wang2019sample},
we
consider transferring the architectures learned from CIFAR-10
to ImageNet \citep{deng2009imagenet}.
We  follow the mobile setting in
\citep{liu2018darts,zoph2018learning}. The size of input image is
$224\times224$ and the number of multiply-add operations is constrained to be fewer
than $600$M. Other training setups are the same

\begin{wraptable}{r}{7.5cm}
\caption{Transferability of different learned architectures on ImageNet.
Here, \#blocks is the number of blocks inside the cell,
and \#mult-adds is the number of multiply-add operations.\label{tab:imagenet}}

\centering{}
\resizebox{0.5\textwidth}{!}{
    \begin{tabular}{cccccc} \hline
    & & & & \multicolumn{2}{c}{error (\%)}\tabularnewline
    & \#blocks & \#mult-adds & \#params & top-1 & top-5 \tabularnewline \hline
    LaNet & 7 & 570 M  &5.1 M & 25.0 & 7.7 \tabularnewline\hline
    NASNet-A & 5 & 564 M & 5.3 M &  26.0 & 8.4 \tabularnewline
    NASNet-B & 5 & 488 M & 5.3 M &  27.2 & 8.7\tabularnewline
    NASNet-C & 5 & 558 M & 4.9 M &  27.5 & 9.0 \tabularnewline
    AmoebaNet-A & 5 & 555 M & 5.1 M &  25.5 & 8.0  \tabularnewline
    AmoebaNet-B & 5 & 555 M &5.3 M &  26.0 & 8.5      \tabularnewline
    AmoebaNet-C & 5 & 570 M &6.4 M &  24.3 & 7.6  \tabularnewline\hline
    PNASNet-5 & 4 & 588 M & 5.1 M & 25.8 & 8.1 \tabularnewline
    DARTS & 4 & 574 M & 4.7 M & 26.7 & 8.7  \tabularnewline
    BayesNAS & 4 & 440 M & 4.0 M &  26.5 & 8.9 \tabularnewline
    BONAS-B & 4 & 500 M & 4.5 M & 24.8 & 7.7 \tabularnewline
    BONAS-C & 4 & 557 M & 5.1 M & 24.6 & 7.5  \tabularnewline
    BONAS-D & 4 & 532 M & 4.8 M & 25.4 & 8.0  \tabularnewline
    \hline
    \end{tabular}}
\vspace{-2mm}
\end{wraptable}
as in
\citep{liu2018darts}.
Since BONAS-A is not competitive on CIFAR-10 compared with other baselines in Section~\ref{subsec:Open-domain-Multi-objective},
we only fully train
BONAS-B/C/D
on ImageNet.

Results are shown in Table~\ref{tab:imagenet} (results of the
baselines are from the corresponding papers).
As can be seen,
in the search space with $4$ blocks (as used by BONAS), the transferred architecture
found by BONAS outperforms
the others.
It's remarkable that BONAS-C achieves a top-$1$ error of $24.6\%$ and a top-$5$ error of $7.5\%$ on ImageNet.
This transferred architecture remains competitive even when compared with baselines using different numbers of blocks in the
search space.

\subsection{Ablation Study}
\label{sec:ablation}

\begin{figure}[t]
\centering
\subfigure[Different BONAS variants.\label{fig:abl}]{\includegraphics[width=0.32\columnwidth]{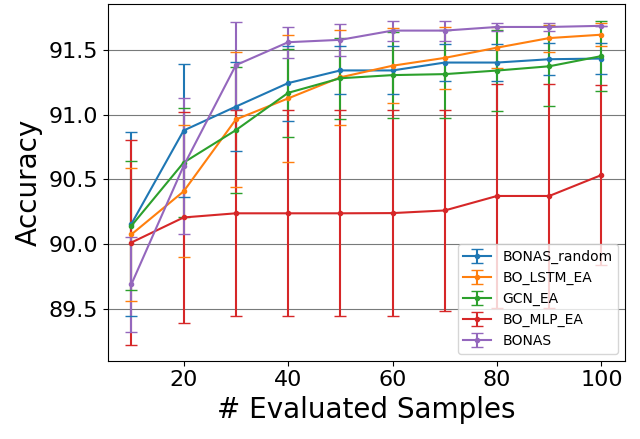}}
\subfigure[Weighted loss vs square loss.\label{fig:loss}]{\includegraphics[width=0.32\columnwidth]{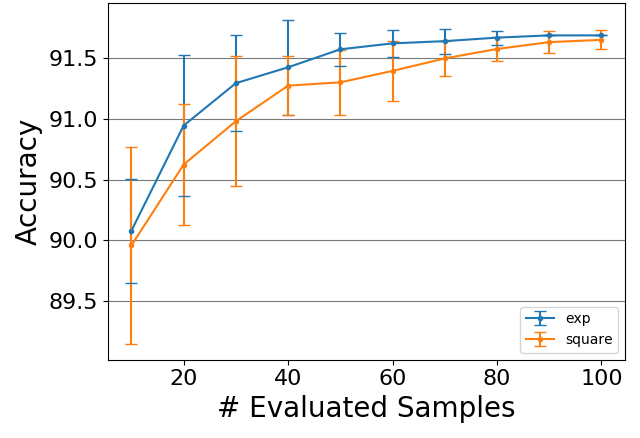}}
\subfigure[Different embedding sizes.\label{fig:gcn_emb}]{\includegraphics[width=0.32\columnwidth]{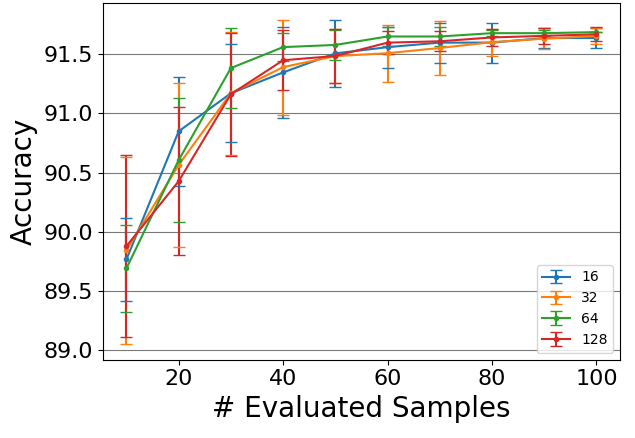}}
\caption{Ablation study on NAS-Bench-201.}
\label{fig:ablation}
\end{figure}

In this section,
we perform ablation study on NAS-Bench-201. The experiment settings are the same as in Section~\ref{subsec:single}.
To investigate the effect of different components of the proposed model, we study the following BONAS variants: (i) BONAS\_random, which replaces EA sampling with random sampling; (ii)
BO\_LSTM\_EA, which replaces the GCN predictor by LSTM;
(iii) BO\_MLP\_EA, which replaces the GCN predictor by MLP;
(iv) GCN\_EA, which removes Bayesian sigmoid regression and uses the
GCN output directly
as selection score.
Results are shown in Figure~\ref{fig:abl}. As can be seen,
BONAS outperforms the various variants.

Next, we compare the proposed weighted loss in (\ref{eq:wl})
with traditional square loss.
Experimental results on NAS-Bench-201 are shown in Figure \ref{fig:loss}.
As can be seen, the use of the proposed loss improves search efficiency by paying
different emphasis on different models.

Finally, to verify the robustness of the BONAS model, we also investigate the influence of embedding size
($\{16, 32, 64, 128\}$). As can be seen from Figure~\ref{fig:gcn_emb}, the
performance is robust to the GCN embedding size.

\section{Conclusion}
In this paper, we proposed BONAS, a sample-based NAS method combined with
weight-sharing paradigm for use with BO.
In the search phase, we use GCN with the Bayesian sigmoid regressor as
BO's
surrogate model
to search for top-performing candidate architectures.
As for query, we adapt the weight-sharing mechanism to query a batch of promising candidate architectures together.
BONAS accelerates sample-based NAS methods and has robust results, thus
bridging the gap between sample-based and one-shot NAS methods. Experiments on
closed-domain search demonstrate its efficiency compared with other sample-based algorithms.
As for open-domain search, we validate BONAS in the NASNet search space, and the
obtained architecture achieves
a top-$1$ error of
$2.43\%$
on CIFAR-10 in $10$ GPU
days.

\section*{Broader Impact}
Neural Architecture Search (NAS) is a powerful framework, and
widely used in the industry to automatically search for models with good
performance. However, the large number of architecture samples required and the
consequent heavy computation are key obstacles for many researchers and small
businesses. NAS also introduces environmental issues
that cannot be overlooked. As pointed out in \cite{strubell2019energy}, the
$\text{CO}_2$ emission
from a NAS process can be comparable to that from
5 cars' lifetime. With the proposed approach, the above-mentioned issues can be alleviated without compromising the final model's performance.

BONAS provides insights to future
NAS research and industrial applications. It allows researchers and
businesses with limited compute to conduct NAS experiments.
This new NAS algorithm is also expected to be more energy-efficient and
environmentally friendly.

\bibliography{neurips_2020}
\bibliographystyle{plain}

\clearpage

\begin{appendices}

\section{Proofs}
\label{app:approx}
\begin{prop} \label{thm:2}
Let
$\mathcal{N}(x|\mu, \sigma^2)$ be the normal distribution with mean $\mu$ and variance $\sigma^2$.
For any given $\alpha$ and $\beta$,
\begin{eqnarray}
\int\Phi(\alpha(x+\beta))\mathcal{N}(x|\mu, \sigma^2)dx =
\Phi\left(\frac{\alpha(\mu+\beta)}{(1+\alpha^2\sigma^2)^{1/2}} \right),
\label{eq:prop}
\end{eqnarray}
where
$\Phi(x)=\int_{-\infty}^x\mathcal{N}(z|0, 1)dz$
is the cumulative distribution function of the standard normal distribution.
\end{prop}
\begin{proof}
Let $z=(x-\mu)/\sigma$, we have
\begin{align*}
    y(\mu,\sigma)
    & = \int\Phi(\alpha(x+\beta))\mathcal{N}(x|\mu, \sigma^2)dx \\
    & = \int\Phi(\alpha(\mu+\sigma z+\beta))\frac{1}{(2\pi\sigma^2)^{1/2}}\exp\{-\frac{1}{2}z^2\}\sigma dz\\
    & = \int\Phi(\alpha(\mu+\sigma z+\beta))\frac{1}{(2\pi)^{1/2}}\exp\{-\frac{1}{2}z^2\} dz.
\end{align*}
Take the derivative of $y$ with respect to $\mu$,
\begin{align*}
    & \frac{\partial y(\mu, \sigma)}{\partial \mu}
    = \frac{\alpha}{2\pi} \int \exp\{-\frac{1}{2}z^2-\frac{1}{2}\alpha^2(\mu+\sigma z+\beta)^2\}dz \\
    & = \frac{\alpha}{2\pi} \int \exp\{-\frac{1}{2}z^2-\frac{1}{2}\alpha^2(\mu^2+\sigma^2 z^2+\beta^2+2\mu\sigma z+2\mu\beta+2\sigma z\beta)\}dz \\
    & = \frac{\alpha}{2\pi} \int \exp\{-\frac{1}{2}(1+\alpha^2\sigma^2)(z^2+\frac{2\alpha^2\sigma(\mu+\beta)}{1+\alpha^2\sigma^2}z+\frac{\alpha^2(\mu^2+\beta^2+2\mu\beta)}{1+\alpha^2\sigma^2})\}dz \\
    & = \frac{\alpha}{2\pi} \int \exp\{-\frac{1}{2}(1+\alpha^2\sigma^2)((z+\frac{\alpha^2\sigma(\mu+\beta)}{1+\alpha^2\sigma^2})^2-\frac{\alpha^4\sigma^2(\mu+\beta)^2}{(1+\alpha^2\sigma^2)^2}+\frac{\alpha^2(\mu+\beta)^2}{1+\alpha^2\sigma^2})\}dz \\
    & = \frac{\alpha}{2\pi} \int \exp\{-\frac{1}{2}(1+\alpha^2\sigma^2)(z+\frac{\alpha^2\sigma(\mu+\beta)}{1+\alpha^2\sigma^2})^2+\frac{1}{2}\frac{\alpha^4\sigma^2(\mu+\beta)^2}{1+\alpha^2\sigma^2}-\frac{1}{2}\alpha^2(\mu+\beta)^2\}dz \\
    & = \frac{\alpha}{2\pi} \int \exp\{-\frac{1}{2}(1+\alpha^2\sigma^2)(z+\frac{\alpha^2\sigma(\mu+\beta)}{1+\alpha^2\sigma^2})^2-\frac{1}{2}\frac{\alpha^2(\mu+\beta)^2}{1+\alpha^2\sigma^2}\}dz \\
    & = \frac{\alpha}{2\pi} \exp\{-\frac{1}{2}\frac{\alpha^2(\mu+\beta)^2}{1+\alpha^2\sigma^2}\} \int \exp\{-\frac{1}{2}(1+\alpha^2\sigma^2)(z+\frac{\alpha^2\sigma(\mu+\beta)}{1+\alpha^2\sigma^2})^2\}dz \\
    & = \frac{1}{(2\pi)^{1/2}}\frac{\alpha}{(1+\alpha^2\sigma^2)^{1/2}}\exp\{-\frac{1}{2}\frac{\alpha^2(\mu+\beta)^2}{1+\alpha^2\sigma^2}\}.
\end{align*}
Similarly, take the derivative of $y$ with respect to $\sigma$,
\begin{align*}
    & \frac{\partial y(\mu, \sigma)}{\partial \sigma}
    = \frac{\alpha}{2\pi} \int  \exp\{-\frac{1}{2}z^2-\frac{1}{2}\alpha^2(\mu+\sigma z+\beta)^2\}zdz \\
    & = \frac{\alpha}{2\pi} \exp\{-\frac{1}{2}\frac{\alpha^2(\mu+\beta)^2}{1+\alpha^2\sigma^2}\}\int \exp\{-\frac{1}{2}(1+\alpha^2\sigma^2)(z + \frac{\alpha^2\sigma(\mu+\beta)}{1+\alpha^2\sigma^2})^2\}zdz \\
    & = -\frac{1}{(2\pi)^{1/2}}\frac{\alpha^3\sigma(\mu+\beta)}{(1+\alpha^2\sigma^2)^{3/2}}\exp\{-\frac{1}{2}\frac{\alpha^2(\mu+\beta)^2}{1+\alpha^2\sigma^2}\}.
\end{align*}

Note that
\begin{eqnarray*}
\frac{\partial \Phi(\frac{\alpha(\mu+\beta)}{(1+\alpha^2\sigma^2)^{1/2}})}{\partial
\mu} & = & \frac{1}{(2\pi)^{1/2}}\frac{\alpha}{(1+\alpha^2\sigma^2)^{1/2}}\exp\{-\frac{1}{2}\frac{\alpha^2(\mu+\beta)^2}{1+\alpha^2\sigma^2}\},\\
\frac{\partial \Phi(\frac{\alpha(\mu+\beta)}{(1+\alpha^2\sigma^2)^{1/2}})}{\partial
\sigma} & = & - \frac{1}{(2\pi)^{1/2}}\frac{\alpha^3\sigma(\mu+\beta)}{(1+\alpha^2\sigma^2)^{3/2}}\exp\{-\frac{1}{2}\frac{\alpha^2(\mu+\beta)^2}{1+\alpha^2\sigma^2}\}.
\end{eqnarray*}
Thus,
\[
 \int\Phi(\alpha(x+\beta))\mathcal{N}(x|\mu, \sigma^2)dx =
 \Phi\left(\frac{\alpha(\mu+\beta)}{(1+\alpha^2\sigma^2)^{1/2}} \right)+C,
\]
for some constant $C$.
When $\alpha=0$,
\[
y(\mu, \sigma) = \int \Phi(0)\mathcal{N}(z|0,1)dz=\frac{1}{2}=\Phi(0),
\]
where Eq.~(\ref{eq:prop}) always holds.
When $\alpha \neq 0$, consider the case where $\mu=-\beta, \sigma=\frac{1}{\alpha}$,
\begin{align*}
y(-\beta, \frac{1}{\alpha}) & = \int \Phi(z)\mathcal{N}(z|0,1)dz \\
& = \int (\Phi(z)-\frac{1}{2})\mathcal{N}(z|0,1)dz + \int \frac{1}{2}\mathcal{N}(z|0,1)dz \\
& = \int \frac{1}{2}\mathcal{N}(z|0,1)dz \\
& = \frac{1}{2}=\Phi(\frac{\alpha(\mu+\beta)}{(1+\alpha^2\sigma^2)^{1/2}}|_{\mu=-\beta, \sigma=\frac{1}{\alpha}}),
\end{align*}
which means $C=0$.
\[
\implies \int\Phi(\alpha(x+\beta))\mathcal{N}(x|\mu, \sigma^2)dx = \Phi(\frac{\alpha(\mu+\beta)}{(1+\alpha^2\sigma^2)^{1/2}}).
\]
\end{proof}

In the following, we align
the function
$\text{sigmoid}(x)$
with
$\Phi(\lambda x)$ (where
$\Phi$ is as defined in Proposition~\ref{thm:1})
such that $\text{sigmoid}(x) \approx \Phi(\lambda x)$.
Obviously, these two functions have the same maxima, minima, and center (at $x=0$).
Thus, we only need to align their derivatives at
$x=0$. Now,
\begin{eqnarray*}
\frac{\partial \text{sigmoid}(x)}{\partial x}|_{x=0} & = &
e^{-x}(1+e^{-x})^{-2}|_{x=0}=\frac{1}{4},\\
\frac{\partial \Phi(\lambda x)}{\partial x}|_{x=0} & = & \frac{\lambda
}{(2\pi)^{1/2}}\exp\{-\frac{1}{2}(\lambda x)^2\}|_{x=0}= \frac{\lambda
}{(2\pi)^{1/2}}.
\end{eqnarray*}
This
implies $\lambda^2 = \frac{\pi}{8}$.

\begin{figure}[ht]
\centering
\subfigure[$\text{sigmoid}(x)$.]
{\includegraphics[width=0.24\columnwidth]{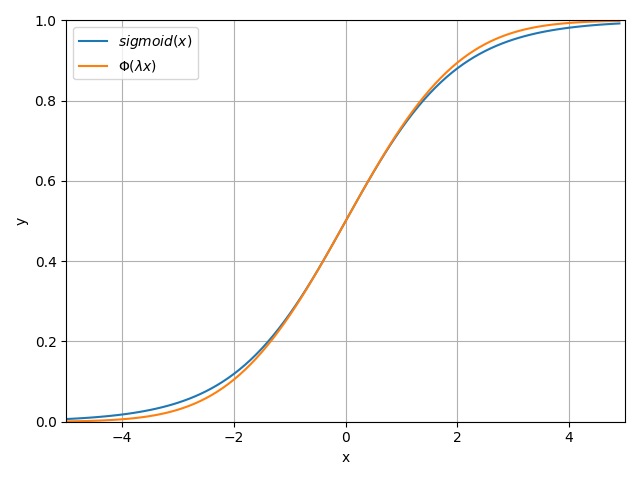}}
\subfigure[$(\text{sigmoid}(x))^2$.]
{\includegraphics[width=0.24\columnwidth]{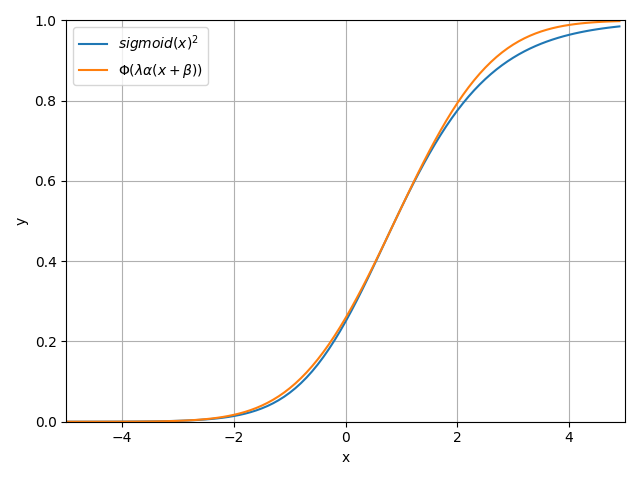}}
\caption{Approximations of $\text{sigmoid}(x)$ and $(\text{sigmoid}(x))^2$.}
\label{fig:sigmoid}
\end{figure}

Similarly, we also align $(\text{sigmoid}(x))^2$ with
$\Phi(\lambda\alpha (x+\beta))$, for some
appropriate $\alpha$ and $\beta$.
Again, note that
both functions have the same maxima  and minima.
The center
of $\Phi(\lambda\alpha (x+\beta))$ is at $(-\beta, 1/2)$.
For alignment, we consider the point when $(\text{sigmoid}(x))^2=1/2$ as its center point, where $x=\log(\sqrt{2}+1)$.
It is easy to see that $\beta=-\log(\sqrt{2}+1)$.
As for the derivative at this center,
\begin{eqnarray*}
\frac{\partial (\text{sigmoid}(x))^2}{\partial x}|_{x=-\beta} & = &
2e^{-x}(1+e^{-x})^{-3}|_{x=-\beta}=(2-\sqrt{2})/2, \\
\frac{\partial \Phi(\lambda \alpha(x+\beta))}{\partial x}|_{x=-\beta} & = & \frac{\lambda\alpha }{(2\pi)^{1/2}}\exp\{-\frac{1}{2}(\lambda \alpha(x+\beta))^2\}|_{x=-\beta}= \frac{\lambda\alpha }{(2\pi)^{1/2}},
\end{eqnarray*}
which
implies $\alpha = 4-2\sqrt{2}$.
Illustrations of the approximations are shown in Figure~\ref{fig:sigmoid}.

Now, using Proposition~\ref{thm:1} and the above approximations, we have
\begin{eqnarray*}
    E[t] & = & \int \frac{1}{\sigma\sqrt{2\pi}}\frac{1}{1-t}\exp(-\frac{(logit(t)-\mu)}{2\sigma^2})dt\\
    & = & \int \text{sigmoid}(x)\mathcal{N}(x|\mu, \sigma^2)dx \simeq \int \Phi(\lambda x)\mathcal{N}(x|\mu, \sigma^2)dx \\
    & = & \Phi(\frac{\lambda\mu}{\sqrt{1+\lambda^2\sigma^2}})\\
    & \simeq & \text{sigmoid}(\frac{\mu}{\sqrt{1+\lambda^2\sigma^2}}),\\
E[t^2] & = & \int \frac{1}{\sigma\sqrt{2\pi}}\frac{t}{1-t}\exp(-\frac{(logit(t)-\mu)}{2\sigma^2})dt \\
    & = & \int \text{sigmoid}(x)^2\mathcal{N}(x|\mu, \sigma^2)dx \\
& \simeq & \int \Phi(\lambda\alpha (x+\beta))\mathcal{N}(x|\mu, \sigma^2)dx \\
    & = & \Phi(\frac{\lambda\alpha(\mu+\beta)}{\sqrt{1+\lambda^2\alpha^2\sigma^2}})
	 \\
& \simeq & \text{sigmoid}(\frac{\alpha(\mu+\beta)}{\sqrt{1+\lambda^2\alpha^2\sigma^2}}), \\
var[t] & = & E[t^2]-E[t]^2 \\
& \simeq & \text{sigmoid}(\frac{\alpha(\mu+\beta)}{\sqrt{1+\lambda^2\alpha^2\sigma^2}}) - (\text{sigmoid}(\frac{\mu}{\sqrt{1+\lambda^2\sigma^2}}))^2
\end{eqnarray*}

\section{Data set}
\label{app:dataset}

\subsection{LSTM-12K Data Set}

We randomly sampled
12K cell structures from the same search space as used in
\citep{pham2018efficient}. The data set consists of 9000 architectures with
7-node cells and 3000 architectures with 8-node cells. There are 4 choices of
operations: ReLU, Sigmoid, Tanh, Identity.
Each architecture is trained for 10 epochs on the PTB data set
\citep{marcus1993building}. Other training setups are the same as
\citep{pham2018efficient}. Specifically, we use SGD with a learning rate
of $20.0$ to train our LSTM models and clip the norm of the gradient
at $0.25$. Besides, we also adapt three same regularization techniques: (i) an
$\ell_2$-regularizer with weight decay parameter $10^{-7}$; (ii)
dropout \cite{gal2016theoretically} with a rate of $0.4$; (iii) tying of
the word embeddings and softmax weights \cite{inan2017tying}.
The models' cell structures, numbers of parameters and perplexities are recorded. This data set can be used to test the efficiency of NAS algorithms before applying them in the open domain.

\subsection{NASNet Search Space}

We follow the search space setting of DARTS \citep{liu2018darts},
in which the architecture is obtained by stacking the learned cell. Each cell
consists of $4$ blocks, two inputs (outputs of the previous cell and
previous previous cell), and one output.
Each intermediate block contains two inputs and one output as follows:
\[
x^{(i)} = o^{(i,j)}(x^{j}) + o^{(i,k)}(x^{k}),
\]
where $x^{(i)}$ is the block output, and $x^{(j)}, x^{(k)}$ are any two
predecessors. There are $7$ types of allowed operations: $3\times3$ and $5\times5$ separable convolutions, $3\times3$
and $5\times5$ dilated separable convolutions, $3\times3$ max pooling,
$3\times3$ average pooling and identity.

Similar to
\citep{liu2018progressive}, we apply the same cell architecture
for both ``normal'' and ``reduction'' layers. In the proposed
GCN predictor, each operation is treated as a node, and each data flow as an edge.

To train
the architecture, we use the same setting as in \cite{liu2018darts}. We use
momentum SGD (with learning rate $0.025$ (anneal cosine strategy), momentum
$0.9$, and weight decay $3\times10^{-4}$).

\section{Illustration of Efficient Estimation}
\label{app:estimation}

To
demonstrate efficiency of the proposed estimation scheme using weight-sharing,
Figure~\ref{fig:vis}
shows the search progress of BONAS on the open domain search in Section~\ref{subsec:Open-domain-Multi-objective}.
Each point in the figure represents a selected architecture.
For each given number of samples searched,
a Gaussian kernel density estimator is fitted on the accuracy distribution of the
selected architectures.
The color corresponds to the corresponding
probability density function value.
As can be seen,
when very few architectures are searched,
the surrogate model cannot estimate the architecture accuracy well, and
the accuracy distribution of the selected models is diffuse.
With more and more samples, the GCN and BSR can perform the
accuracy estimation better.
After around $2000$ samples, most of the candidate models selected by BONAS have
high estimated accuracies.

For sub-networks that are sampled in a particular search iteration,
Figure~\ref{fig:diff}
compares their actual accuracies (obtained by full training) with the
estimated accuracies obtained by the proposed method
(Section~\ref{subsec:weight-share}) and
standard weight-sharing (which constructs the
super-network by using all models in the search space).
As can be seen, the proposed weight-sharing among a smaller
number of promising models can achieve higher correlation.

\begin{minipage}[b]{0.48\textwidth}
\includegraphics[width=1\columnwidth]{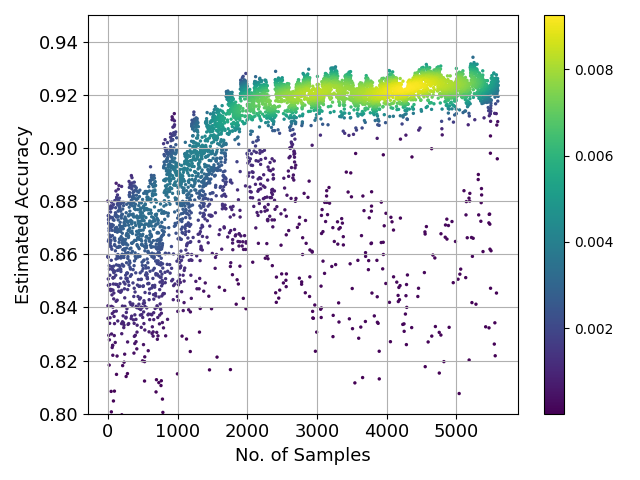}
\captionof{figure}{Visualization of BONAS's search progress on open-domain search.}
\label{fig:vis}
\end{minipage}
\hfill
\begin{minipage}[b]{0.48\textwidth}
\includegraphics[width=1\columnwidth]{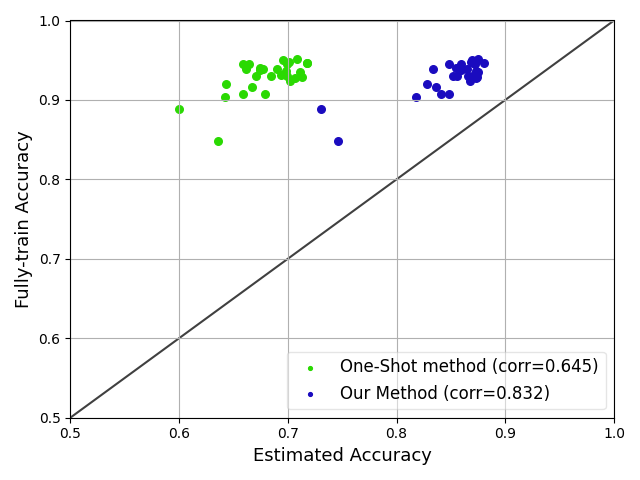}
\captionof{figure}{Actual accuracy versus estimated accuracies obtained by
the proposed method and one-shot NAS.}
\label{fig:diff}
\end{minipage}


\clearpage
\section{Example Architectures Obtained \label{app:Best-Found-Models}}

\begin{figure}[htb]
\centering
\subfigure[BONAS-A.]{\includegraphics[width=0.4\columnwidth]{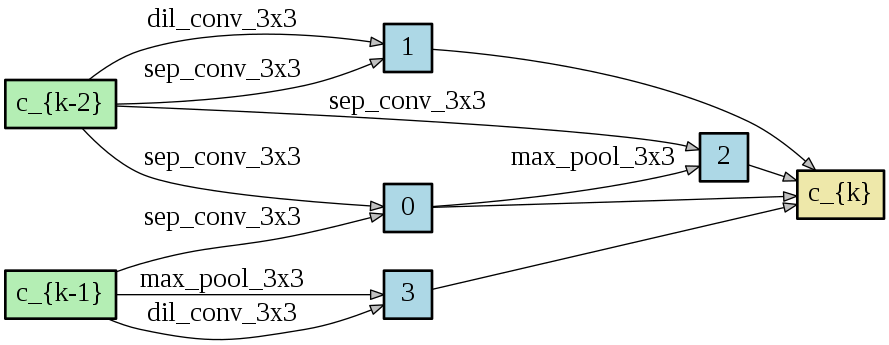}}
\subfigure[BONAS-B.]{\includegraphics[width=0.4\columnwidth]{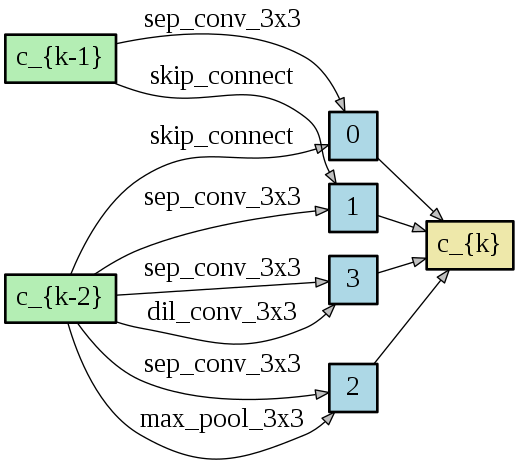}}
\subfigure[BONAS-C.]{\includegraphics[width=0.4\columnwidth]{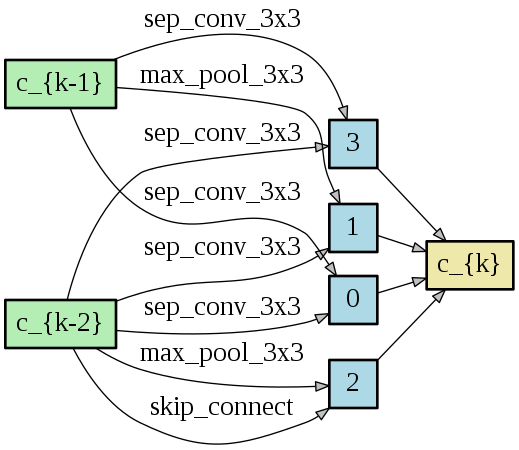}}
\subfigure[BONAS-D.]{\includegraphics[width=0.4\columnwidth]{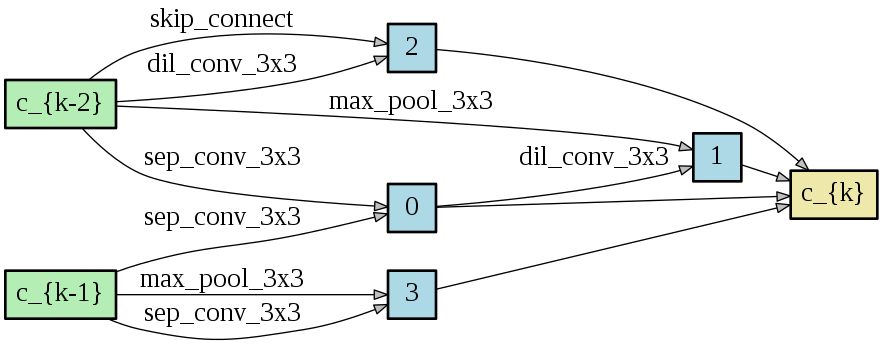}}
\caption{Example models obtained by BONAS in the NASNet search space.}
\label{fig:found}
\end{figure}

Figure \ref{fig:found} shows some example architectures that are obtained
by the proposed method
from open-domain search on the NASNet search space
(Section~\ref{subsec:Open-domain-Multi-objective}).

\end{appendices}

\end{document}